\documentclass{article}

\usepackage[preprint]{neurips_2025}

\usepackage{booktabs}       %
\usepackage{longtable}
\usepackage{multirow}

\usepackage[utf8]{inputenc} %
\usepackage[T1]{fontenc}    %
\usepackage{hyperref}       %
\usepackage{url}            %
\usepackage{amsfonts}       %
\usepackage{nicefrac}       %
\usepackage{microtype}      %
\usepackage[table]{xcolor}
\usepackage{soul}           %

\usepackage[pdftex]{graphicx}
\usepackage{listliketab}
\usepackage{hyperref}
\hypersetup{
    colorlinks=true,
    linkcolor=blue,
    filecolor=magenta,
    urlcolor=blue,
    citecolor=blue
    }
\usepackage{amsmath}
\usepackage{ulem} %
\usepackage{cleveref} %
\usepackage{todonotes}
\usepackage[section]{placeins}
\usepackage{enumitem}

\usepackage{caption}
\usepackage{amsmath}

\usepackage[compact]{titlesec}

\titlespacing{\section}{0pt}{1.0ex plus 0.2ex minus 0.2ex}{0.6ex}
\titlespacing{\subsection}{0pt}{0.8ex plus 0.2ex minus 0.2ex}{0.4ex}
\titlespacing{\subsubsection}{0pt}{0.6ex plus 0.2ex minus 0.2ex}{0.3ex}

\newcommand{\pnpl}{\includegraphics[height=2.2ex]{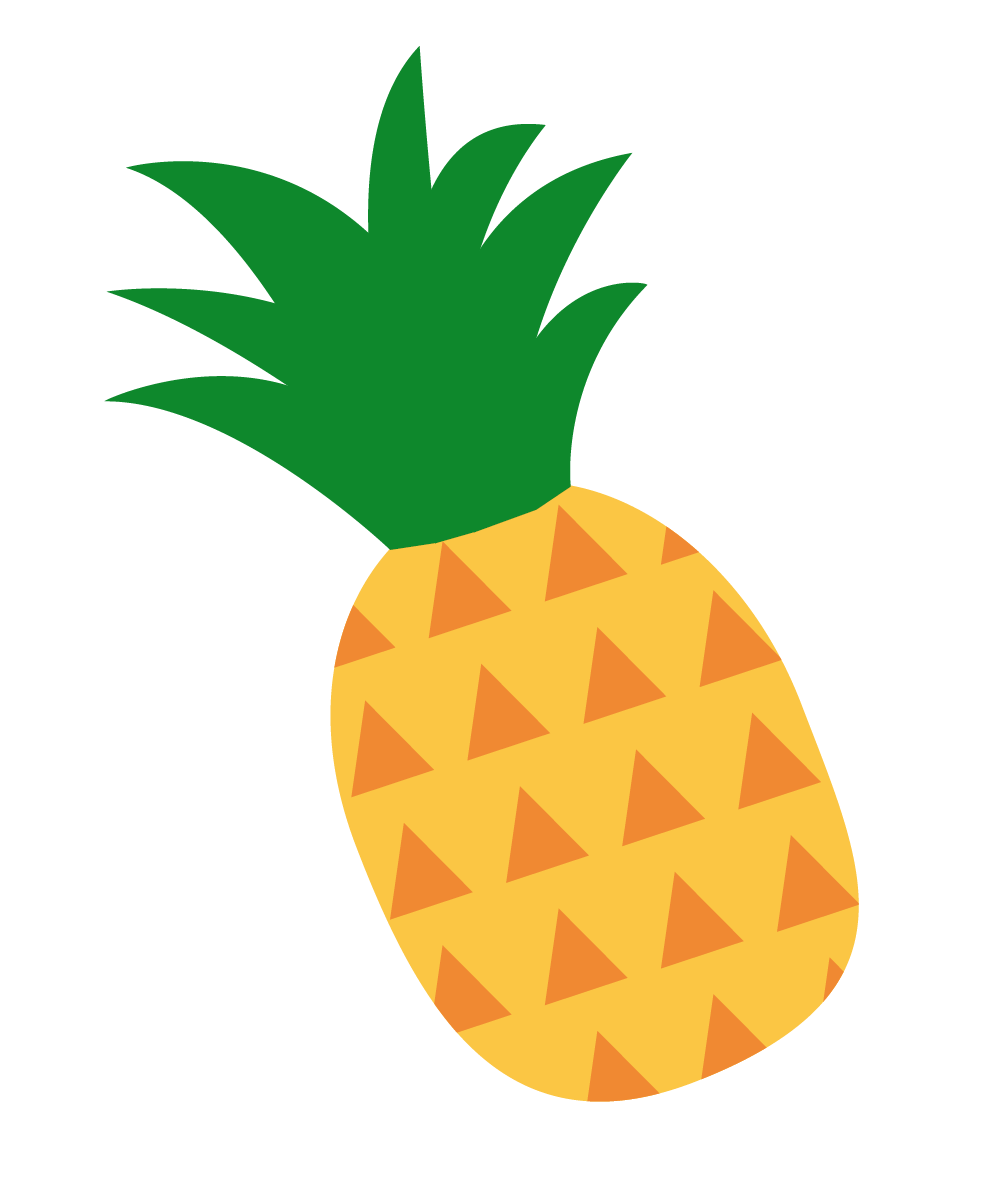}%
}

\usepackage{minted}

\setminted{
  frame=none,
  bgcolor=gray!20,
  fontsize=\footnotesize,
  linenos,
  breaklines=true
} %

\usepackage{subcaption} %

\usepackage{float}
\floatstyle{plaintop}
\restylefloat{table}
\usepackage[tableposition=top]{caption}

\title{The 2026 PNPL Competition: Word Classification and Efficient Cross-Subject Generalisation in LibriBrain100}

\author{%
Francesco Mantegna$^{1,*}$ 
\quad Gereon Elvers$^{1,*}$
\quad Dulhan Jayalath$^{1,*}$ 
\quad Gilad Landau$^{1}$ 
\AND Tasha Kim$^{1}$
\quad Miran \"{O}zdogan$^{1}$ 
\quad Luisa Kurth$^{1}$ 
\quad Teyun Kwon$^{1}$ 
\quad SungJun Cho$^{1,2}$ 
\AND Benjamin Ballyk$^{1}$ 
\quad Alex Fung$^{1,3}$ 
\quad Anna Greer$^{1}$ 
\quad Pratik Somaiya$^{1}$ 
\quad Christian Herff$^{4}$
\AND Yorguin Mantilla Ramos$^{5}$
\quad Hamza Abdelhedi$^{5}$
\quad Karim Jerbi$^{5,6}$
\quad Greg Farquhar$^{7}$
\AND Brendan Shillingford$^{7}$
\quad Mark Woolrich$^{2}$ 
\quad Oiwi Parker Jones$^{1}$\\ \\
$^1$PNPL\includegraphics[height=2.2ex]{logos/PNPL_logos_final_Pineapple.png}, Department of Engineering Science, University of Oxford, UK \\ \quad 
$^2$OHBA, Oxford Centre for Integrative Neuroimaging, University of Oxford, UK \\
$^3$FMRIB, Oxford Centre for Integrative Neuroimaging, University of Oxford, UK \\
$^4$Maastricht University, The Netherlands \quad %
$^5$UNIQUE, Université de Montréal, Canada \\
$^6$Mila--Quebec AI Institute, Canada \quad
$^7$Google DeepMind, UK \\
\\ $^*$Joint first authors \\
\\ \texttt{\{francesco, gereon, dulhan, oiwi\}@robots.ox.ac.uk}
}

\begin{document}

\maketitle
\begin{abstract}
    
The ambition of the 2025 PNPL competition~\citep{landau2025competition} was to launch a multi-year curriculum for non-invasive speech decoding. Designed to progress from foundational tasks toward the linguistic complexity required for a practical brain-computer interface (BCI), it set the stage with speech detection and phoneme classification tasks. Winning submissions reached F1-macro  scores of 95.6\% and 73.6\% on the respective tasks~\citep{elvers2026benchmarking}, highly significant advances. %
This success was built on the LibriBrain dataset~\citep{ozdogan2025libribrain}, the largest within-subject MEG dataset recorded at the time with ${\sim}50$ hours of data for one subject. However, while within-subject scale drives strong decoding performance, a practical BCI must generalise to new users from minutes of data, not hours. %
 
The 2026 PNPL competition responds to this challenge with LibriBrain100~\citep{mantegna2026libribrain100}, an extended LibriBrain dataset with 32 additional subjects (${\sim}40$ minutes each) plus even more within-subject data (${\sim}80$ hours). Advancing the curriculum of tasks to focus on word classification, two complementary tracks are presented in this competition: the \textit{Deep} track targets within-subject word classification at scale, aiming at the best possible performance; the \textit{Broad} track targets cross-subject generalisation, progressively reducing the amount of subject-specific fine-tuning data from ${\sim}40$ to ${\sim}20$ to ${\sim}10$ minutes, a duration that falls within a clinically feasible range and brings us a step closer to a non-invasive BCI capable of restoring communication to people living with profound paralysis.

\end{abstract}

\section{Competition description}
\begin{figure}[tbp]
    \centering
\includegraphics[width=\linewidth]{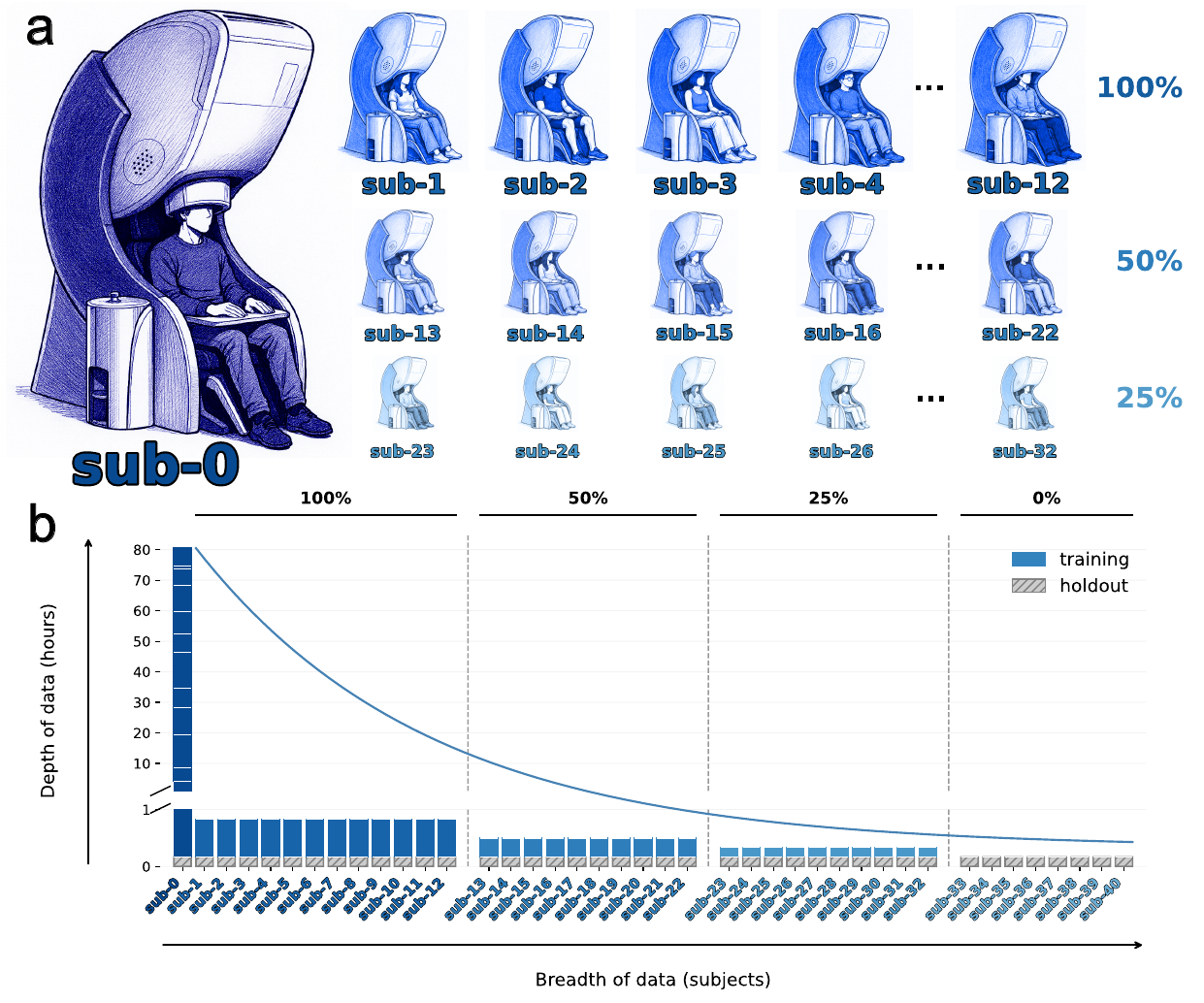}
    \caption{\textbf{Subject-wise data splits}. (\textbf{a}) Schematic illustration of the dataset. A single extensively recorded subject is shown alongside other subjects with progressively reduced amounts of available training data. (\textbf{b}) Quantitative representation of the data splits. 
    A broken y-axis is used to accommodate the large difference between sub-0 and the other subjects, while preserving the visibility of differences in training data proportions. Vertical dashed lines separate the subject groups, and labels indicate the proportion of training data available within each group. A decreasing curve illustrates the overall reduction in available training data across subjects.}
    \label{fig:broad+deep}
\vspace{-1.5em}
\end{figure}
\subsection{Background and impact}\label{sec:background}

Invasive brain-computer interfaces (BCIs) for speech have advanced at a remarkable pace. Since the first demonstration of connected-speech decoding from surgically implanted electrodes in a paralysed individual~\citep{moses2021nejm}, vocabulary sizes have rapidly grown from 50 words to over 125,000 words~\citep{willett2023high} while word-error rates (WERs) have shrunk to 2.5\%~\citep{card2024nejm}---a superhuman score well below the $\sim$5--10\% reported for human transcription on standard automatic speech recognition (ASR) benchmarks~\citep{xiong2016parity}. 
Yet invasive BCIs face fundamental barriers to widespread deployment: brain surgery carries inherent risks, per-patient data collection does not easily scale, and the populations who might benefit most are often those least able to undergo elective neurosurgery; non-invasive alternatives are therefore the real prize. 

Among non-invasive neuroimaging modalities, magnetoencephalography (MEG) stands out for its combination of millisecond temporal resolution and typical spatial precision around 5--10~mm~\citep{hamalainen1993meg}---with the capacity to go even lower to 2--4 mm~\citep{barratt2018meg2mm}---putting it in the range of invasive modalities like ECoG but without the surgical risks. Recent years have seen genuine advances in MEG-based speech decoding~\citep{defossez2023, dascoli2025natcomm, jayalath2025bbl-icml}, but progress has been harder to interpret and harder to compare across studies than in the invasive setting. A key reason is the lack of shared infrastructure: common datasets, fixed evaluation splits, standard metrics, and baseline implementations that allow the community to build cumulatively on each other's work.

The PNPL competition series is motivated by the view that closing this gap will require not only better models, but better community infrastructure. The 2025 PNPL competition~\citep{landau2025competition} was designed with this goal in mind. Building on the LibriBrain dataset~\citep{ozdogan2025libribrain}---the largest within-subject MEG dataset recorded at the time, with ${\sim}50$ hours of data from a single participant---it introduced common benchmarks for speech detection and phoneme classification, together with a Python library for automatic data download and loading, standardised train/validation/test splits for replicability, reference models, a public leaderboard, tutorial code, and a dedicated competition website. These tasks were deliberately chosen to be relatively simple: speech detection reduces to binary classification over time samples, while phoneme classification is a well-defined supervised learning problem with manageable output structure. This was part of a broader curricular design---by starting with accessible tasks, we aimed to lower the barrier to entry for researchers from machine learning who might not yet have the domain knowledge to confidently navigate human brain data, while laying the groundwork for more ambitious forms of neural speech decoding. Winning submissions reached F1-macro scores of 95.6\% and 73.6\% on the respective tasks, driving significant advances in the field~\citep{elvers2026benchmarking}. 
Over the summer of 2025, the 2025 PNPL competition attracted 155 registered teams from 15 countries and generated 6,041 submissions. By April 2026, the LibriBrain dataset had reached over 16,000 downloads  %
--- indicating a clear demand for the materials we provided, even after the 2025 competition ended.

The 2026 PNPL competition represents a logical next step, with significantly more data and larger ambitions. 
A central reason this next step is now feasible is the scale of the expanded LibriBrain100 dataset \citep{mantegna2026libribrain100}. Compared with existing public MEG speech datasets, LibriBrain100 occupies a distinctive position: it combines substantially greater within-subject depth with a larger multi-subject cohort, supporting both high-performance within-subject modelling and systematic evaluation of cross-subject generalisation (visualised in Appendix \ref{app:dataset-comparison}). 
This combination of depth and breadth is essential for the two-track design of the competition: the Deep track exploits the unusually large amount of data available for a single subject, while the Broad track tests whether models trained across subjects can adapt efficiently to new individuals.
The task of word classification is a natural progression in the curriculum: it remains sufficiently structured to support clear benchmarking and broad participation, but it is substantially closer to the longer-term goal of full brain-to-text (B2T)~\citep{herff2015}, which is analogous to speech-to-text but with brain signals as inputs. Whereas speech detection and phoneme classification primarily target lower-level acoustic structure in the signal, word classification brings lexical and semantic information into scope and begins to probe whether models can recover meaningful units of speech from MEG data. As a stepping stone toward a working non-invasive B2T, it is well-motivated: word classification was a key component of the pipeline in the first invasive BCI for a paralysed individual~\citep{moses2021nejm}, and remains a useful intermediate target given the apparent difficulty of non-invasive B2T~\citep{jo2024eegtotext} --- though we have finally begun to see the first WER scores for non-invasive speech B2T that are significantly better than chance~\citep{jayalath2025unlocking}. 
Although Brain2Qwerty models report interesting results for decoding sentences from MEG acquired while subjects overtly type on a keyboard~\citep{levy2026brain2qwerty,zhang2026brain2qwerty2}, we view this as distinct from speech BCIs: there is currently no clear path for typing-based decoding to be used by patients unable to move their fingers. 

Word classification has recently attracted growing attention in the non-invasive decoding literature~\citep{dascoli2025natcomm, jayalath2025unlocking, jayalath2025meg-xl}. Yet without standard evaluation protocols, results remain difficult to compare across studies. A key source of incomparability is the choice of target vocabulary. Two studies that report the same evaluation metric on the most frequent few hundred words in their respective datasets may appear comparable~\citep[e.g.,][]{dascoli2025natcomm, jayalath2025unlocking}, but this is complicated if the target words differ or if their frequencies vary between datasets~\citep{jayalath2026ovmi}. The choice of words also affects what can be communicated. For all of these reasons, it is important to report the target vocabulary explicitly.
Official competition rankings are evaluated on a 50-word \textit{competition vocabulary} designed to leverage high-frequency words in the training corpora and to support expressive communication (Appendix~\ref{sec:retrieval_vocab}). 

To make it easier to compare competition results with prior results in the literature, we further track performance on a second 50-word vocabulary, the \textit{Moses 50 vocabulary}, which has been used many times in invasive studies~\citep[e.g.,][]{moses2021nejm, willett2023high}. Finally, we report an information-theoretic measure designed to compare results with different target vocabularies: open-vocabulary mutual information (OVMI)~\citep{jayalath2026ovmi}. OVMI accounts for both decoding performance and vocabulary coverage (Section~\ref{sec:metrics}).

\subsection{Novelty}\label{sec:novelty}

The 2025 PNPL competition was the first dedicated to language decoding from non-invasive brain data~\citep{landau2025competition}. Building on its success, the 2026 competition introduces several elements that are new both to the PNPL series and to the broader field.

\textbf{Word classification as a standardised competition task.} 
Although word classification has recently attracted attention in the non-invasive decoding literature~\citep{dascoli2025natcomm, jayalath2025unlocking, jayalath2025meg-xl}, standardised evaluation is harder here than for tasks with fixed symbol sets such as phoneme classification. Unlike phonemes, words form an effectively open set, and the choice of vocabulary---not merely its size---affects measured performance~\citep{jayalath2026ovmi}. Custom vocabularies, such as the most-frequent $K$ words in a corpus~\citep{dascoli2025natcomm}, maximise training examples and reported accuracy scores. On the other hand, results for custom vocabularies are not comparable across studies and high-frequency words are often dominated by function words of limited communicative utility. Standardised cross-study vocabularies such as the \textit{Moses~50}~\citep{moses2021nejm} enable comparisons (including with invasive BCIs), but may be poorly attested in a given corpus. OVMI~\citep{jayalath2026ovmi} addresses both concerns, jointly accounting for decoding accuracy and corpus coverage. Because they achieve different goals, we use all three methods of evaluation (Section~\ref{sec:metrics}) --- a practice we recommend to the field. The primary metric for both the leaderboard and final ranking is the custom, dataset-tailored top-10 balanced accuracy.

\textbf{Cross-subject generalisation as a competition target.} The 2025 competition and the LibriBrain dataset focused exclusively on a single participant. Notably, all recent state-of-the-art invasive systems have likewise been trained and evaluated on a single individual~\citep{moses2021nejm, metzger2023neuroprosthesis, willett2023high, card2024nejm}, meaning cross-subject generalisation has received little attention in the field as a whole. The 2026 competition is the first to explicitly target it, introducing a dedicated \textit{Broad} track that evaluates how well models can adapt to new individuals with progressively limited data.
This mirrors the constraints of clinical deployment and provides a novel evaluation paradigm for data-efficient generalisation. 
Uniquely, the Broad track also includes holdout subjects for whom no subject-specific training data is released at all, enabling evaluation of zero-shot cross-subject generalisation --- the most clinically relevant regime.

\paragraph{Contrast with invasive brain-to-text competitions.}
Building on high-performance but invasive speech neuroprosthesis systems~\citep{willett2023high, card2024nejm}, the Brain-to-Text Benchmark~'24 and its successor competition target the decoding of full transcripts from intracortical recordings~\citep{willett2024benchmark}. 
Our goal is to build the corresponding non-invasive benchmark for MEG. Since open-vocabulary B2T from speech-related neural activity remains out of reach for current non-invasive systems~\citep[though see][]{jayalath2025unlocking}, we instead take a curricular approach and focus on word classification. This gives the community a concrete intermediate task on which to measure progress in representation learning, cross-subject generalisation, and decoding under the constraints of non-invasive neural data.

\textbf{LibriBrain100 as competition resource.} The competition is the first to be built around a large-scale, multi-subject MEG dataset. LibriBrain100~\citep{mantegna2026libribrain100} combines the deepest within-subject MEG dataset recorded to date (${\sim}80$ hours) with data from 32 additional subjects, enabling both within-subject and cross-subject benchmarking within a single competition.

\subsection{Data}\label{sec:data}

For this year's competition, we primarily use the LibriBrain100 dataset~\citep{mantegna2026libribrain100}, which comprises non-invasive MEG recordings from 33~subjects listening to naturalistic speech (stimuli details in Appendix \ref{app:corpora}). Subject~0---the single participant from the original LibriBrain dataset~\citep{ozdogan2025libribrain}---is extended to ${\sim}80$~hours of recordings, the deepest within-subject MEG dataset recorded to date. For an additional 32~subjects, ${\sim}40$~minutes of MEG data were collected, but for the competition we have initially released tiered amounts: the full ${\sim}40$~minutes for 12~subjects, ${\sim}20$~minutes for 10, and ${\sim}10$~minutes for a further~10---a regime within the range of what will often be clinically feasible (Figure~\ref{fig:broad+deep}). The motivation for this is to challenge competition participants to predict with decreasing amounts of subject-specific data. For a final group of 8~subjects not included in LibriBrain100 (subjects~33--40), there are no subject-specific training data, requiring zero-shot cross-subject generalisation: the most demanding but clinically useful regime. We will release the full data for subjects 1--32  after the competition.

The LibriBrain100 data come with standard train, validation, and test splits for reproducibility. Evaluation in the competition is conducted exclusively on an additional competition \textit{holdout} set, which 
is drawn from sources not revealed to participants in advance. For the competition, we release the MEG recordings for this holdout set but withhold the labels. Submissions consist of predictions on these recordings, uploaded to our evaluation platform. The holdout is divided into two disjoint partitions: one used to update the public leaderboard throughout the competition, and one reserved for final ranking of submissions. This design mirrors the 2025 competition and is intended to prevent overfitting to the holdout distribution while still giving participants timely feedback on their progress.

MEG data and labels are saved in HDF5 and TSV formats, respectively. The data are also available on Hugging Face in both serialised (HDF5) and raw (FIF) formats, across two repositories: \texttt{libribrain} contains the original LibriBrain recordings, while \texttt{libribrain2} contains the additional data comprising LibriBrain100. Note that raw FIF files have not been preprocessed and include head movement artefacts and other noise; they are also substantially larger than the serialised versions. %

To make interacting with the dataset easy, we provide an updated Python library that automatically downloads and loads data for PyTorch. It can be installed with a single command: \texttt{pip install pnpl}. We recommend using the library to load serialised data, as it handles downloads automatically and retrieves only the partitions requested (Appendix~\ref{sec:code_samples}); no knowledge of underlying repo structures is 
required.

\subsection{Tasks and applications}\label{sec:tasks}
 
The task in this competition is word classification from non-invasive brain recordings. Given a window of MEG data $x \in \mathbb{R}^{306 \times T}$ (sensors $\times$ time samples), the goal is to predict the word $y \in \mathcal{V}$ being heard by the participant, where $\mathcal{V}$ is a fixed retrieval vocabulary \citep[see][]{mantegna2026libribrain100}. We use a custom 50-word vocabulary designed for reliable evaluation on LibriBrain100 (see Appendix~\ref{sec:retrieval_vocab}). Relevant neural information for this task may include both phonetic representations (reflecting acoustic and articulatory processing in auditory and motor cortices) and lexical semantic representations which cover most of the cortex~\citep{huth2016}. Given the distributed nature of semantic representations in particular, MEG's whole-brain coverage is a distinct advantage over surgically implanted arrays, which 
sample only a limited (surgically accessed) cortical region. This makes word classification a richer decoding target than either speech detection or phoneme classification alone.
 
Word classification occupies an important position in the curriculum of non-invasive speech decoding. It is more structured than open-vocabulary brain-to-text, which has only recently yielded word error rates better than chance from non-invasive recordings~\citep{jayalath2025unlocking}, making it a reliable and reproducible benchmark. At the same time, it goes substantially beyond the tasks of the 2025 competition~\citep{landau2025competition}: speech detection and phoneme classification target sub-lexical structure, but word classification directly probes the recovery of meaningful linguistic units. It is also practically motivated. Word classification, combined with speech detection and a language model, formed the core pipeline of the first invasive speech BCI for a paralysed individual~\citep{moses2021nejm}. Solving word classification from non-invasive data, together with the speech detection foundation laid in 2025, would therefore constitute a meaningful step toward a non-invasive analogue.
 
The competition hosts two complementary word classification tasks that run concurrently:
 
(a) $Deep$ track: This track evaluates word classification in a single, densely-sampled subject. Participants may train on any data they wish.
The goal is the best possible decoding performance on the densely-sampled individual (Subject 0), to push the limits of non-invasive word classification.
 
(b) $Broad$ track: This track evaluates cross-subject generalisation. The challenge is to produce accurate word classification predictions for individuals with limited subject-specific data: for 12 subjects, ${\sim}40$ minutes of subject-specific data is available; for 10 subjects, ${\sim}20$ minutes; and for a further 10 subjects, ${\sim}10$ minutes. The ${\sim}10$ minute regime is of particular interest for clinical applications, where extended patient data collection is often infeasible. 
Holdout data is provided for a final set of 8 subjects with no released subject-specific training data, enabling evaluation of zero-shot generalisation.
To perform well on this track, submissions must generalise across all data regimes---motivating the emphasis on \textit{data-efficiency} 
in the competition title.

\subsection{Metrics}\label{sec:metrics}

To measure success in the word classification tasks we use the Top-10 Balanced Accuracy
with a fixed retrieval vocabulary of $K = 50$:
\[
\mathrm{BAcc@10} = \frac{1}{K}\sum_{k=1}^{K}\mathrm{Recall@10}_k.
\]
This metric has been used many times for decoding words in the recent decoding literature,
with retrieval vocabularies ranging from $K=50$ to $K=250$~\citep{dascoli2025natcomm,
jayalath2025meg-xl}. For the competition, we track both a custom 50-word vocabulary and the Moses 50 vocabulary (Appendix~\ref{sec:retrieval_vocab}). 
$\mathrm{Recall@10}_k$ denotes the Top-10 recall for
class $k$:
\[
\mathrm{Recall@10}_k = \frac{1}{N_k}\sum_{i:\,y_i=k}
\mathbb{I}\!\left[y_i \in \{\hat{y}_{i,1}, \ldots, \hat{y}_{i,10}\}\right],
\]
where $y_i$ is the true label, $\hat{y}_{i,r}$ the class with the $r$-th
highest predicted score, and $N_k$ the number of evaluation examples with true label $k$.
We use a balanced metric so that each word contributes equally, regardless of how frequently
it appears in the evaluation set~\citep[see, e.g.,][]{tholke2023bacc}. In practice, $\mathrm{BAcc@10}$ scores range from 0 to 1
and can be reported as percentages. For a vocabulary of $K=50$ words, uniform random
guessing without replacement yields a $\mathrm{BAcc@10}$ of 20\% in expectation, since the
correct class has probability $10/50$ of appearing in a random top-10 set. A score of
100\% means that the correct word is always included in the model's top-10 predictions.
A limitation of this metric is that it does not distinguish between cases where the correct word is ranked
first or tenth; therefore, we also compute the Top-1 Balanced Accuracy
($\mathrm{BAcc@1}$) to help break ties.

Finally, as an auxiliary metric, we report Open-Vocabulary Mutual
Information~(OVMI)~\citep{jayalath2026ovmi}. OVMI is designed to support comparison across decoders with different target vocabularies. It is an information-theoretic quantity that
measures the mutual information between a user's intent and the output of a decoding model relative to a reference communication distribution. 
Concretely, $\mathrm{OVMI}(S) = C(S)\,I(X;Y \mid X \in S)$, where $C(S)$ is the lexical coverage of the retrieval vocabulary $S$ under a reference distribution $p$ and $I(X;Y \mid X \in S)$ is the in-vocabulary MI between the intended word $X$ and the
decoder output $Y$. 
Conceptually, $p$ acts like a prior over the words a user is likely to intend. For the competition, we use SUBTLEX-UK as $p$~\citep{vanHeuven2014SubtlexUKAN} and report OVMI in bits per word perceived under this distribution. OVMI is $0$ under uniform random guessing; higher OVMI scores are better.

\subsection{Baselines, code, and material provided}\label{sec:baselines}
 
To provide participants with baseline scores that are representative of the state-of-the-art and straightforward to implement and improve upon, we provide two reference models---one for each track.
 
The \textit{Deep} track uses an implementation of the supervised word decoding model of~\citet{dascoli2025natcomm} as reference. This model achieves strong within-subject word classification performance and provides a competitive baseline for participants targeting the best possible performance on Subject 0.
 
The \textit{Broad} track uses MEG-XL~\citep{jayalath2025meg-xl} as a reference model. MEG-XL is a self-supervised foundation model pre-trained on ${\sim}300$ hours of MEG from 800 subjects, intended to be fine-tuned for word classification. Crucially, it is designed to generalise well with small amounts of subject-specific fine-tuning data, where it outperforms the \citet{dascoli2025natcomm} model, making it particularly well-suited to the data-efficient cross-subject generalisation setting of the \textit{Broad} track. Code and checkpoints are available through GitHub\footnote{\url{https://github.com/neural-processing-lab/MEG-XL}}, providing participants with a competitive starting point that they can directly improve upon. The choice of reference models reflects recent advances in the field: with large amounts of within-subject data, supervised models outperform self-supervised ones~\citep{jayalath2025meg-xl}, while with limited per-subject data, pre-trained priors of self-supervised models like MEG-XL confer a substantial advantage~\citep{mantegna2026libribrain100}.
 
For our baseline results, we train or fine-tune each reference model following the procedure described in their respective papers and report results on the competition test splits (Table~\ref{tab:scores}). For the Deep track, we use all the available training data for Subject 0. For the Broad track, across Subjects 1--32, we use half of the data from the Sherlock 1 Session 11 recording for fine-tuning and the other half for validation. We suspect joint training on Subject 0 and Subjects 1--32 will improve generalisation on the latter subjects further, but leave this and other avenues of exploration to competition participants.

\begin{table}[]
\centering
\small
\renewcommand{\arraystretch}{0.85}
\begin{tabular}{lc>{\columncolor{gray!10}}cc@{\hspace{1.5em}}ccc}
\toprule
    & \multicolumn{3}{c}{\textbf{Competition Vocabulary}}
    & \multicolumn{3}{c}{\textbf{Moses Vocabulary}} \\
\cmidrule(lr){2-4}
\cmidrule(lr){5-7}
    \textbf{Method}
    & $\mathrm{BAcc@1}$ & $\mathrm{BAcc@10}$ & OVMI
    & $\mathrm{BAcc@1}$ & $\mathrm{BAcc@10}$ & OVMI \\
\midrule
\textit{Deep track} & & \\
    \textbf{d'Ascoli (reference)} & $\mathbf{25.60}\%$ & $\mathbf{73.23}\%$ & $\mathbf{.220}$ & $\mathbf{38.84}\%$ & $\mathbf{83.60
    }\%$ & $\mathbf{.157}$ \\
    \textbf{Random chance} & $2.00\%$ & $20.00\%$ & $.000$ & $2.00\%$ & $20.00\%$ & $.000$ \\
\midrule
\textit{Broad track} & & \\
    \textbf{MEG-XL (reference)} & $\mathbf{6.20\%}$ & $\mathbf{42.96}\%$ & $\mathbf{.014}$ & $\mathbf{10.21}\%$ & $\mathbf{52.23}\%$ & $\mathbf{.011}$ \\
    \textbf{d'Ascoli} & $5.14\%$ & $33.13\%$ & $.009$ & $6.10\%$ & $49.56\%$ & $.006$ \\
    \textbf{Random chance} & $2.00\%$ & $20.00\%$ & $.000$ & $2.00\%$ & $20.00\%$ & $.000$ \\
\bottomrule
\end{tabular}
    \caption{\textbf{Scores for reference models and random baselines.} The column in \protect\textcolor{gray!15}{\protect\rule{1.1ex}{1.1ex}} (grey) highlights the competition evaluation metric ($\mathrm{BAcc@10}$ on the competition vocabulary). All scores are on each track's test data from the checkpoint with the best validation performance on the competition metric.}
    \label{tab:scores}
\vspace{-1.5em}
\end{table}
 
As an entry point to the library and competition, we provide three (3) tutorial notebooks in the form of interactive Google Colab notebooks, optimised for usage with a free GPU:
\begin{enumerate}[itemsep=0pt, topsep=0pt]
    \item \emph{First tutorial}: Data loading, introduction to LibriBrain100 and word classification task. Contains code for fine-tuning the reference model on Subject 0.
    \item \emph{Second tutorial}: Loading subject-specific data, fine-tuning the reference model for cross-subject generalisation, more details on the \textit{Broad} track.
    \item \emph{Third tutorial}: Prediction generation on the holdout data, submission to the leaderboard.
\end{enumerate}

\subsection{Competition website}\label{sec:website}

The competition website\footnote{\url{https://libribrain.com/editions/2026/}} serves as a central hub, aggregating the starter kit, documentation, tutorial notebooks, leaderboards, and links to the Kaggle submission platform that accepts prediction files in CSV format. To maximise ease of onboarding, all tutorials are provided as interactive Colab notebooks that run in the browser, requiring no local installation beyond: \texttt{pip install pnpl}.

\vspace{-0.5em}

\section{Organisation}
\subsection{Protocol}\label{sec:protocol}
 
Participating is designed to be simple and accessible. To participate, one can install the \texttt{pnpl} Python library (\texttt{pip install pnpl}), which automatically downloads the dataset and provides 
a standard PyTorch \texttt{DataLoader}. Once data is loaded, participants can develop solutions and generate predictions on the \textit{holdout} data. The \texttt{pnpl} library can be used to write these predictions to a CSV file, which can then be uploaded to the submission platform (Appendix \ref{sec:code_samples}). Solutions are evaluated automatically according to the metrics in Section~\ref{sec:metrics}, and rankings updated continuously on the public leaderboard.
 
Both the $Deep$ and $Broad$ tracks run concurrently for three months (15 July -- 15 October). Participants can submit to either or both tracks with progress reflected on the appropriate leaderboard. 
\vspace{-0.5em}

\subsection{Rules and engagement}\label{sec:rules}

\textbf{Goal.} The competition aims to be open and accessible to a broad community of researchers. This includes participants that may have no prior experience in analysing neural data. Our rules are intended to encourage open participation, while protecting integrity of the evaluation process.

\textbf{Competition rules.}
\begin{enumerate}[itemsep=0pt, topsep=0.5pt]
    \item Participation is open to everyone. 
    Domain-specific knowledge or specialised hardware is \emph{not} required for fully participating in the competition. 

    \item \textbf{Competition tracks}. The competition is organised around the LibriBrain100 dataset and consists of two distinct tracks, both targeting word classification using balanced accuracy over a fixed vocabulary: (a) $Deep$ track: word classification within a single, deeply-sampled subject, and (b) $Broad$ track: word classification across many held-out subjects. 

    Note, the two tracks are ranked separately, with (i) a leaderboard and (ii) prizes separately tracked for each. Participants may submit to either or both tracks. 

    \item \textbf{Permitted materials}. Participants may use any training data for either track. This includes (i) external, publicly available datasets, (ii) self-collected or synthetically-generated data, or (iii) pre-trained models available from public repositories on Hugging Face  (\url{https://huggingface.co}) or GitHub (\url{https://github.com/}). Participants must independently ensure their data use complies with the appropriate licensing and ethical requirements.

    \item \textbf{Winners}. Final rankings are determined against an independent \emph{holdout dataset} to prevent data leakage. We discourage any attempt to infer held-out labels outside of the official evaluation procedure. The organisers reserve the right to make final decisions on edge cases.

    \item \textbf{Prize distribution}. The highest-ranked three (3) submissions in each track are to be considered for prizes, provided they beat the performance of reference model baselines described in Section~\ref{sec:baselines}. Each team is eligible to win \emph{at most} one prize across the competition. If the same team places in the prize-winning positions on both tracks, they will be listed on all relevant leaderboards, but prizes for the additional track pass to the next eligible team. In the unlikely event of a tie, the prize will be split between the tied teams. %

    \item The organisers will contact all prize contenders for additional verification of their submissions. Participants may be asked to share the training code, model checkpoints, or descriptions of their approach. If neither correctness of the approach nor compliance with the competition rules can be verified, the submission may be excluded from prize eligibility.

    \item The organising team encourages participants to share their code in the spirit of open science, including direct contributions (e.g., data loaders, reusable baselines, documentation) to the \texttt{pnpl} GitHub repository (\url{https://github.com/neural-processing-lab/pnpl}).
\end{enumerate}

\textbf{Communication and community.}
To facilitate open and real-time communication,
the competition uses the same Discord server established during last year's competition\footnote{\url{https://libribrain.com/links/discord}}.

\subsection{Schedule and readiness}\label{sec:schedule}
\vspace{-0.5em}
All essential competition materials are online. For prizes, the equivalent of \$5,000 USD has been raised.

\begin{itemize}[itemsep=0pt, topsep=0pt]

    \item \textbf{15 July, 2026}: Submissions to both tracks officially open. All necessary materials are released, including documentation and unlabelled evaluation data for each track.

    \item \textbf{15 July -- 15 October, 2026}: Competition progress is tracked on the public leaderboards. The organising team provides support through the competition website, GitHub, and public Discord server. Participants may submit to either track during the competition period.

    \item \textbf{15 October, 2026}: Submissions to both tracks officially close.

    \item \textbf{15 October -- 30 November, 2026}: The organising team reviews and verifies the highest-ranking submissions. Contenders for all  prizes will be contacted to further confirm submission details. 

    \item \textbf{1 December, 2026}: The confirmed winners are announced through the competition website. The organising team independently conducts post-competition analysis of the results. 
    Winners will be connected with sponsors to coordinate prizes. 

\end{itemize}

\subsection{Competition promotion and incentives}\label{sec:promotion}
\vspace{-0.5em}
We are promoting the competition through social networks and academic mailing lists (e.g., NeurIPS affinity groups). We launched a dedicated website to share information, resources, and updates regarding the competition. 
This includes live leaderboards. 
During the competition, the organising team plans to release blog posts about the tasks and data to foster discussion.

\section{Resources}
\vspace{-0.5em} 
The \texttt{pnpl} Python library is available to download the data and integrate it seamlessly into standard deep learning frameworks. Reference model code and checkpoints are publicly available on GitHub. 
The tutorial code runs on Google Colab in the browser, providing some GPU access to get started.

\newpage 
\section*{Organising team}

\textbf{Francesco Mantegna} is a Postdoctoral Researcher in PNPL{\pnpl}, Department of Engineering Science, University of Oxford. He received his PhD from NYU under the supervision of David Poeppel. 

\textbf{Gereon Elvers} is an Encode:~AI for Science Fellow in PNPL{\pnpl}, University of Oxford. He was previously a master's student, visiting PNPL from the Technical University of Munich (TUM).

\textbf{Dulhan Jayalath} is a DPhil student in Machine Learning at the University of Oxford, supervised by Oiwi Parker Jones as part of PNPL{\pnpl} and funded by an Amazon Web Services (AWS) studentship as part of the EPSRC Centre for Doctoral Training in Autonomous Intelligent Machines and Systems (AIMS). 

\textbf{Gilad Landau} is a DPhil student in Engineering and member of PNPL{\pnpl}, supervised by Oiwi Parker Jones at the University of Oxford. 

\textbf{Tasha Kim} is a DPhil student in Engineering and member of PNPL{\pnpl}, supervised by Oiwi Parker Jones at the University of Oxford. 

\textbf{Miran Özdogan} is a DPhil student in Computer Science and member of PNPL{\pnpl}, supervised by Oiwi Parker Jones and Michael Bronstein, at the University of Oxford. 

\textbf{Luisa Kurth} is a DPhil student in PNPL{\pnpl} and the EPSRC Centre for Doctoral Training in Autonomous Intelligent Machines and Systems (AIMS), University of Oxford.

\textbf{Teyun Kwon} is a DPhil student in PNPL{\pnpl} and the EPSRC Centre for Doctoral Training in Autonomous Intelligent Machines and Systems (AIMS), University of Oxford.

\textbf{SungJun Cho} is a DPhil student in Neuroscience supervised by Mark Woolrich and Oiwi Parker Jones at the University of Oxford. 

\textbf{Benjamin Ballyk} is a DPhil student in Engineering and member of PNPL{\pnpl}, supervised by Oiwi Parker Jones at the University of Oxford. 

\textbf{Alex Fung} is a DPhil student in Neuroscience and member of PNPL{\pnpl}, supervised by Alex Green, Saad Jbabdi, and Oiwi Parker Jones at the University of Oxford. 

\textbf{Anna Greer} is a DPhil student in the EPSRC Centre for Doctoral Training in Autonomous Intelligent Machines and Systems (AIMS), University of Oxford. She contributed during a rotation mini-project in PNPL{\pnpl}. 

\textbf{Pratik Somaiya} is a Software Engineer at the Oxford Robotics Institute. 

\textbf{Christian Herff} is an Associate Professor and head of the Neural Interfacing Lab in the Department of Neurosurgery at Maastricht University.

\textbf{Yorguin Mantilla Ramos} is a master's student at the Université de Montréal and Graduate Research Assistant at Mila. 

\textbf{Hamza Abdelhedi} is a Biomedical Engineering PhD student at the Université de Montréal, supervised by Karim Jerbi and specialising in Neuro-AI.

\textbf{Karim Jerbi} is Professor in the Psychology Department of the Université de Montréal and Associate Professor at Mila. He heads UNIQUE, the Quebec-wide Neuro-AI research center, and is also Canada Research Chair in Computational Neuroscience and Cognitive Neuroimaging. 

\textbf{Greg Farquhar} is a Staff Research Scientist at Google DeepMind. %

\textbf{Brendan Shillingford} is a Staff Research Scientist at Google DeepMind. %

\textbf{Mark Woolrich} is Professor of Computational Neuroscience at the University of Oxford. He is Head of Analysis and Associate Director of the Oxford Centre for Human Brain Activity (OHBA).

\textbf{Oiwi Parker Jones} heads the \textit{Parker Jones Neural Processing Lab}
(PNPL\pnpl) in the Department of Engineering Science, University of Oxford. He is also a Fellow in Computer Science at Jesus College Oxford, a Principal Investigator at the Oxford Robotics Institute, and an Honorary Fellow in the Nuffield Department of Clinical Neurosciences.

\section*{Acknowledgments}\label{sec:ack}
Many thanks to Pillar VC for sponsoring this year's competition. 
We gratefully acknowledge the use of the University of Oxford Advanced Research Computing (ARC) facility (\url{http://dx.doi.org/10.5281/zenodo.22558}), Hartree Centre resources, and the NVIDIA Corporation for donating additional GPUs. PNPL is supported by the MRC (MR/X00757X/1), Royal Society (RG$\backslash$R1$\backslash$241267), NSF (2314493), NFRF (NFRFT-2022-00241), SSHRC (895-2023-1022), and ARIA (SCNI-SE01-P004).

\clearpage

\bibliography{references}

@article{herff2015,
    title = {Brain-to-text: decoding spoken phrases from phone representations in the brain},
    author = {Herff, C. and Heger, D. and {de Pesters}, A. and Telaar, D. and Brunner, P. and Schalk, G. and Schultz, T.},
    journal = {Frontiers in Neuroscience},
    volume = {9},
    number = {217},
    pages = {1--11},
    year = {2015}
}

@article{huth2016,
    author = {Huth, Alexander G. and de Heer, Wendy A. and Griffiths, Thomas L. and Theunissen, Frédéric E. and Gallant, Jack L.},
    title = {Natural speech reveals the semantic maps that tile human cerebral cortex},
    year = {2016},
    journal = {Nature},
    volume = {532},
    pages = {453--458}
}

@article{willett2023high,
  title={A high-performance speech neuroprosthesis},
  author={Willett, Francis R and Kunz, Erin M and Fan, Chaofei and Avansino, Donald T and Wilson, Guy H and Choi, Eun Young and Kamdar, Foram and Glasser, Matthew F and Hochberg, Leigh R and Druckmann, Shaul and Shenoy, Krishna V and Henderson, Jaimie M},
  journal={Nature},
  volume={620},
  pages={1031--1036},
  year={2023},
  doi={10.1038/s41586-023-06377-x}
}

@article{card2024nejm,
  author = {Card, Nicholas S. and Wairagkar, Maitreyee and Iacobacci, Carrina and Hou, Xianda and Singer-Clark, Tyler and Willett, Francis R. and Kunz, Erin M. and Fan, Chaofei and Vahdati Nia, Maryam and Deo, Darrel R. and Srinivasan, Aparna and Choi, Eun Young and Glasser, Matthew F. and Hochberg, Leigh R. and Henderson, Jaimie M. and Shahlaie, Kiarash and Stavisky, Sergey D. and Brandman, David M.},
  title = {An accurate and rapidly calibrating speech neuroprosthesis},
  journal = {New England Journal of Medicine},
  volume = {391},
  number = {7},
  pages = {609--618},
  year = {2024}
}

@article{dascoli2025natcomm,
  author  = {d'Ascoli, St{\'e}phane and Bel, Corentin and Rapin, J{\'e}r{\'e}my and Banville, Hubert and Benchetrit, Yohann and Pallier, Christophe and King, Jean-R{\'e}mi},
  title   = {Towards decoding individual words from non-invasive brain recordings},
  journal = {Nature Communications},
  year    = {2025},
  volume  = {16},
  pages   = {10521},
  doi     = {10.1038/s41467-025-65499-0},
  url     = {https://www.nature.com/articles/s41467-025-65499-0},
  month   = nov
}

@article{defossez2023,
  author = {Défossez, Alexandre and Caucheteux, Charlotte and Rapin, Jérémy and Kabeli, Ori and King, Jean-Rémi},
  title = {Decoding speech perception from non-invasive brain recordings},
  journal = {Nature Machine Intelligence},
  volume = {5},
  number = {10},
  pages = {1097--1107},
  year = {2023},
  doi = {10.1038/s42256-023-00714-5},
  url = {https://www.nature.com/articles/s42256-023-00714-5}
}

@article{gwilliams2023megmasc,
  author = {Gwilliams, Laura and Flick, Graham and Marantz, Alec and Pylkkanen, Liina and Poeppel, David and King, Jean-Rémi},
  title = {Introducing {MEG-MASC}: A high-quality magneto-encephalography dataset for evaluating natural speech processing},
  journal = {Scientific Data},
  volume = {10},
  number = {1},
  pages = {862},
  year = {2023},
  doi = {10.1038/s41597-023-02752-5},
  url = {https://www.nature.com/articles/s41597-023-02752-5}
}

@article{jo2024eegtotext,
  author = {Jo, Hyejeong and Yang, Yiqian and Han, Juhyeok and Duan, Yiqun and Xiong, Hui and Lee, Won Hee},
  title = {Are {EEG}-to-text models working?},
  journal = {arXiv preprint},
  year = {2024},
  archivePrefix = {arXiv},
  eprint = {2405.06459},
  url = {https://arxiv.org/abs/2405.06459},
  note = {https://arxiv.org/abs/2405.06459}
}

@article{metzger2023neuroprosthesis,
  author = {Metzger, Sean L. and Littlejohn, Kaylo T. and Silva, Alexander B. and Moses, David A. and Seaton, Margaret P. and Wang, Ran and Dougherty, Maximilian E. and Liu, Jessie R. and Wu, Peter and Berger, Michael A. and Zhuravleva, Inga and Tu-Chan, Adelyn and Ganguly, Karunesh and Anumanchipalli, Gopala K. and Chang, Edward F.},
  title = {A high-performance neuroprosthesis for speech decoding and avatar control},
  journal = {Nature},
  volume = {620},
  pages = {1037--1046},
  year = {2023}
}

@article{moses2021nejm,
  author = {Moses, David A. and Metzger, Sean L. and Liu, Jessie R. and Anumanchipalli, Gopala K. and Makin, Joseph G. and Sun, Pengfei F. and Chartier, Josh and Dougherty, Maximilian E. and Liu, Patricia M. and Abrams, Gary M. and Tu-Chan, Adelyn and Ganguly, Karunesh and Chang, Edward F.},
  title = {Neuroprosthesis for Decoding Speech in a Paralyzed Person with Anarthria},
  journal = {New England Journal of Medicine},
  volume = {385},
  number = {3},
  pages = {217--227},
  year = {2021},
  doi = {10.1056/NEJMoa2027540},
  url = {https://doi.org/10.1056/NEJMoa2027540}
}

@article{tang2023,
  author = {Tang, Jerry and LeBel, Amanda and Jain, Shailee and Huth, Alexander G.},
  title = {Semantic reconstruction of continuous language from non-invasive brain recordings},
  journal = {Nature Neuroscience},
  volume = {26},
  number = {5},
  pages = {858--866},
  year = {2023},
  doi = {10.1038/s41593-023-01304-9},
  url = {https://www.nature.com/articles/s41593-023-01304-9}
}

@article{ozdogan2025libribrain,
  author    = {Özdogan, Miran and Landau, Gilad and Elvers, Gereon and Jayalath, Dulhan and Somaiya, Pratik and Mantegna, Francesco and Woolrich, Mark and Parker Jones, Oiwi},
  title     = {{LibriBrain}: Over 50 Hours of Within-Subject {MEG} to Improve Speech Decoding Methods at Scale},
  year      = {2025},
  journal   = {NeurIPS, Datasets and Benchmarks Track},
}

@article{tholke2023bacc,
  title     = {Class imbalance should not throw you off balance: Choosing the right classifiers and performance metrics for brain decoding with imbalanced data},
  author    = {Thölke, Philipp and Mantilla-Ramos, Yorguin-Jose and Abdelhedi, Hamza and Maschke, Charlotte and Dehgan, Arthur and Harel, Yann and Kemtur, Anirudha and Mekki Berrada, Loubna and Sahraoui, Myriam and Young, Tammy and Bellemare Pépin, Antoine and El Khantour, Clara and Landry, Mathieu and Pascarella, Annalisa and Hadid, Vanessa and Combrisson, Etienne and O'Byrne, Jordan and Jerbi, Karim},
  journal   = {NeuroImage},
  volume    = {277},
  pages     = {120253},
  year      = {2023},
  doi       = {10.1016/j.neuroimage.2023.120253},
}

@article{landau2025competition,
  title        = {The 2025 {PNPL} Competition: Speech Detection and Phoneme Classification in the {LibriBrain} Dataset},
  author       = {Landau, Gilad and Özdogan, Miran and Elvers, Gereon and Mantegna, Francesco and Somaiya, Pratik and Jayalath, Dulhan and Kurth, Luisa and Kwon, Teyun and Shillingford, Brendan and Farquhar, Greg and Jiang, Minqi and Jerbi, Karim and Abdelhedi, Hamza and Mantilla Ramos, Yorguin and Gulcehre, Caglar and Woolrich, Mark and Voets, Natalie and Parker Jones, Oiwi},
  year         = {2025},
  journal         = {NeurIPS, Competition Track},
  note = {arXiv preprint arXiv:2506.10165}
}

@article{barratt2018meg2mm,
  title = {Mapping the topological organisation of beta oscillations in motor cortex using {MEG}},
  author = {Barratt, Eleanor L. and Francis, Susan T. and Morris, Peter G. and Brookes, Matthew J.},
  journal = {NeuroImage},
  volume = {181},
  pages = {831--844},
  year = {2018},
  publisher = {Elsevier},
  doi = {10.1016/j.neuroimage.2018.06.041},
  url = {https://doi.org/10.1016/j.neuroimage.2018.06.041}
}

@misc{garofolo1993timit,
  title={{TIMIT} Acoustic-Phonetic Continuous Speech Corpus},
  author={Garofolo, John S and Lamel, Lori F and Fisher, William M and Fiscus, Jonathan G and Pallett, David S and Dahlgren, Nancy L and Zue, Victor},
  year={1993},
  howpublished={Linguistic Data Consortium},
  note={\url{https://catalog.ldc.upenn.edu/LDC93S1}}
}

@misc{wrench1999mocha_timit,
  author       = {Wrench, Alan},
  title        = {{The MOCHA-TIMIT articulatory database}},
  year         = {1999},
  month        = nov,
  howpublished = {\url{https://www.cstr.ed.ac.uk/research/projects/artic/mocha.html}},
  note         = {Created November 1999; accessed 2026-02-07}
}

@article{armeni2022,
  author       = {Armeni, K. and G{\"u}{\c{c}}l{\"u}, U. and van Gerven, M. and Schoffelen, J.-M.},
  title        = {A 10-hour within-participant magnetoencephalography narrative dataset to test models of language comprehension},
  journal      = {Scientific Data},
  volume       = {9},
  pages        = {278},
  year         = {2022}
}

@article{schoffelen2019mous,
  author = {Schoffelen, Jan-Mathijs and Oostenveld, Robert and Lam, Nietzsche H. L. and Uddén, Julia and Hultén, Annika and Hagoort, Peter},
  title        = {A 204-subject multimodal neuroimaging dataset to study language processing},
  journal      = {Scientific Data},
  volume       = {6},
  number = {17},
  pages        = {1--13},
  year         = {2019},
  month        = apr
}

@article{xiong2016parity,
  title={Achieving Human Parity in Conversational Speech Recognition},
  author={Xiong, Wayne and Droppo, Jasha and Huang, Xuedong and Seide, Frank and Seltzer, Michael and Stolcke, Andreas and Yu, Dong and Zweig, Geoffrey},
  journal={arXiv preprint arXiv:1610.05256},
  year={2016},
  url={https://arxiv.org/abs/1610.05256}
}

@article{willett2024benchmark,
  title     = {{Brain-to-Text Benchmark '24}: Lessons Learned},
  author    = {Willett, Francis R. and Li, Jingyuan and Le, Trung and Fan, Chaofei and Chen, Mingfei and Shlizerman, Eli and Chen, Yue and Zheng, Xin and Okubo, Tatsuo S. and Benster, Tyler and Lee, Hyun Dong and Kounga, Maxwell and Buchanan, E. Kelly and Zoltowski, David and Linderman, Scott W. and Henderson, Jaimie M.},
  journal   = {arXiv preprint arXiv:2412.17227},
  year      = {2024},
  url       = {https://doi.org/10.48550/arXiv.2412.17227},
  eprint    = {2412.17227},
  archivePrefix = {arXiv},
  primaryClass  = {cs.CL}
}

@book{doyle1887study,
  author = {Arthur Conan Doyle},
  title = {A Study in Scarlet},
  year = {1888},
  publisher = {Ward, Lock \& Co},
}

@book{doyle1890sign,
  author = {Arthur Conan Doyle},
  title = {The Sign of the Four},
  year = {1890},
  publisher = {Spencer Blackett},
}

@book{doyle1892adventures,
  author = {Arthur Conan Doyle},
  title = {The Adventures of Sherlock Holmes},
  year = {1892},
  publisher = {George Newnes},
}

@book{doyle1893memoirs,
  author = {Arthur Conan Doyle},
  title = {The Memoirs of Sherlock Holmes},
  year = {1893},
  publisher = {George Newnes},
}

@book{doyle1902hound,
  author = {Arthur Conan Doyle},
  title = {The Hound of the Baskervilles},
  year = {1902},
  publisher = {George Newnes},
}

@book{doyle1905return,
  author = {Arthur Conan Doyle},
  title = {The Return of Sherlock Holmes},
  year = {1905},
  publisher = {George Newnes},
}

@book{doyle1915valley,
  author = {Arthur Conan Doyle},
  title = {The Valley of Fear},
  year = {1915},
  publisher = {George H. Doran Company},
}

@book{doyle1917lastbow,
  author = {Arthur Conan Doyle},
  title = {His Last Bow},
  year = {1917},
  publisher = {John Murray},
}

@book{doyle1927casebook,
  author = {Arthur Conan Doyle},
  title = {The Case-Book of Sherlock Holmes},
  year = {1927},
  publisher = {John Murray},
}

@article{jayalath2025unlocking,
  title={Unlocking Non-Invasive Brain-to-Text},
  author={Jayalath, Dulhan and Landau, Gilad and Parker Jones, Oiwi},
  journal={International Conference on Machine Learning (ICML), Workshop on Generative AI and Biology},
  note={arXiv preprint arXiv:2505.13446},
  year={2025}
}

@article{jayalath2025meg-xl,
  title={{MEG-XL}: Data-Efficient Brain-to-Text via Long-Context Pre-Training},
  author={Jayalath, Dulhan and Parker Jones, Oiwi},
  journal={International Conference on Machine Learning (ICML)},
    series       = {Proceedings of Machine Learning Research},
  publisher    = {{PMLR}},
  year={2026},
  note={arXiv preprint arXiv:2602.02494}
}

@article{jayalath2025bbl-icml,
  author       = {Dulhan Jayalath and
                  Gilad Landau and
                  Brendan Shillingford and
                  Mark W. Woolrich and
                  Oiwi {Parker Jones}},
  title        = {{The Brain's Bitter Lesson}: Scaling Speech Decoding With Self-Supervised
                  Learning},
  journal    = {International Conference on Machine Learning (ICML)},
  publisher    = {{PMLR}},
  year         = {2025},
  note = {arXiv preprint arXiv:2406.04328}
}

@article{mantegna2026libribrain100,
         title={{LibriBrain100}: One Hundred Hours of Broad and Deep {MEG} Data for Neural Speech Decoding at Scale},
         author={Mantegna, Francesco and Jayalath, Dulhan and Elvers, Gereon and Kim, Tasha and Ballyk, Benjamin and Fung, Alex and Cho, SungJun and Kwon, Teyun and Kurth, Luisa and \"Ozdogan, Miran and Landau, Gilad and Somaiya, Pratik and Voets, Natalie and Woolrich, Mark and Parker Jones, Oiwi},
         year={2026},
         journal={arXiv preprint arXiv:2608.25204}
}

@article{elvers2026benchmarking,
  title   = {Benchmarking Non-Invasive Speech {BCIs}: Lessons Learned from the 2025 {PNPL} Competition},
  author  = {Elvers, Gereon and
             Landau, Gilad and
             Mantegna, Francesco and
             {\"O}zdogan, Miran and
             Kim, Tasha and
             Kwon, Teyun and
             Cho, SungJun and
             Ballyk, Benjamin and
             Kurth, Luisa and
             Jayalath, Dulhan and
             Somaiya, Pratik and
             {de Zuazo}, Xabier and
             Shillingford, Brendan and
             Farquhar, Greg and
             Jiang, Minqi and
             Jerbi, Karim and
             Abdelhedi, Hamza and
             {Mantilla Ramos}, Yorguin and
             Gulcehre, Caglar and
             Woolrich, Mark and
             Voets, Natalie and
             {Parker Jones}, Oiwi},
  journal = {Journal of Machine Learning Research},
  year    = {2026},
  note    = {In press},
}

@article{hamalainen1993meg,
  author    = {H{\"a}m{\"a}l{\"a}inen, Matti and Hari, Riitta and Ilmoniemi, Risto J. and Knuutila, Jukka and Lounasmaa, Olli V.},
  title     = {Magnetoencephalography---theory, instrumentation, and applications to noninvasive studies of the working human brain},
  journal   = {Reviews of Modern Physics},
  volume    = {65},
  number    = {2},
  pages     = {413--497},
  year      = {1993},
  publisher = {American Physical Society}
}

@misc{jayalath2026ovmi,
  title={A Common Measure of Communication for Speech Brain--Computer Interfaces},
  author={Jayalath, Dulhan and Ballyk, Benjamin and Parker Jones, Oiwi},
  year={2026},
  note={Manuscript in preparation}
}

@article{vanHeuven2014SubtlexUKAN,
  title={{SUBTLEX-UK}: A New and Improved Word Frequency Database for {British} {English}},
  author={Walter J. B. van Heuven and Paweł Mandera and Emmanuel Keuleers and Marc Brysbaert},
  journal={Quarterly Journal of Experimental Psychology},
  year={2014},
  volume={67},
  pages={1176 - 1190},
}

@article{levy2026brain2qwerty,
  title={Noninvasive decoding of typed sentences from human brain activity},
  author={L{\'e}vy, Jarod and Zhang, Mingfang and Pinet, Svetlana and Rapin, J{\'e}r{\'e}my and Banville, Hubert and d'Ascoli, St{\'e}phane and King, Jean-R{\'e}mi},
  journal={Nature Neuroscience},
  year={2026},
  doi={10.1038/s41593-026-02303-2}
}

@article{zhang2026brain2qwerty2,
  title={Accurate Decoding of Natural Sentences from Non-Invasive Brain Recordings},
  author={Zhang, Mingfang and L{\'e}vy, Jarod and Rommel, Cedric and Rapin, J{\'e}r{\'e}my and Bel, Corentin and Bonnaire, Julie and Nieto, Daniel and Bourdillon, Pierre and Pinet, Svetlana and d'Ascoli, St{\'e}phane and Moreau, Thomas and King, Jean-R{\'e}mi},
  journal={arXiv preprint arXiv:2608.18114},
  year={2026}
}
\bibliographystyle{apalike}

\newpage
\appendix
\section*{Appendices / Supplemental Materials}

\section{Dataset Comparison}
\label{app:dataset-comparison}

\begin{figure}[htbp]
    \centering

    \includegraphics[width=\linewidth]{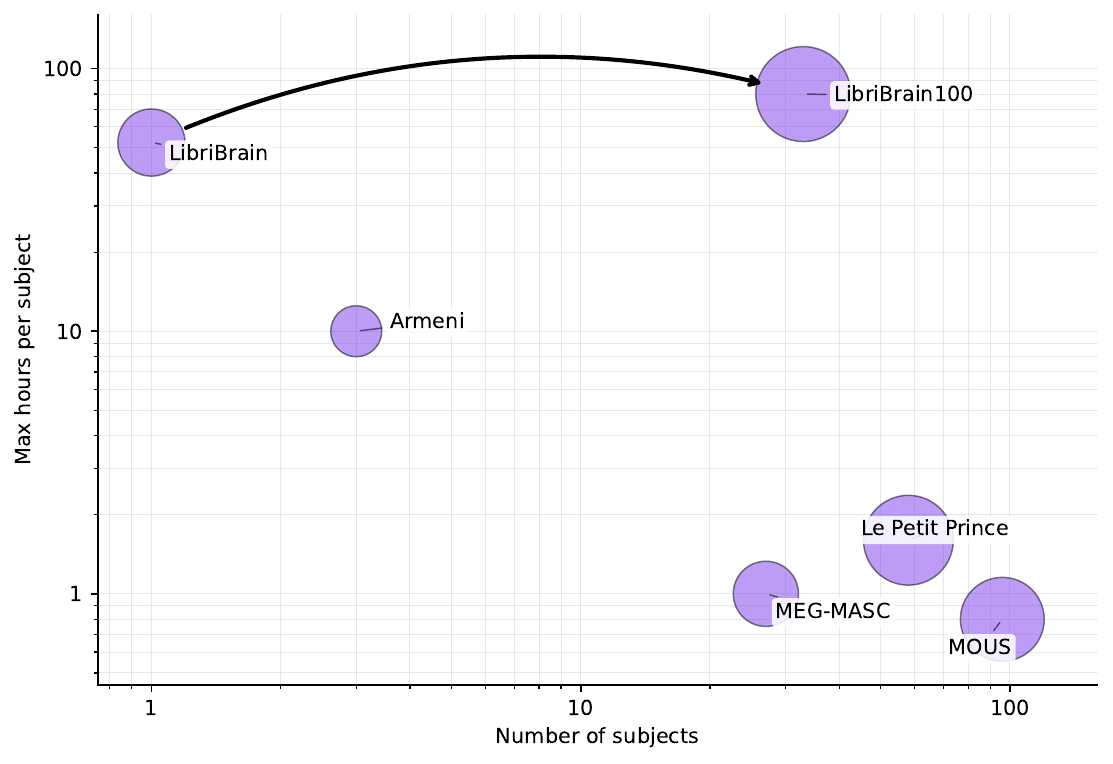}

    \caption{
    \textbf{Dataset comparison.}
    Comparison of public MEG speech datasets by number of subjects and maximum hours per subject. Both axes are logarithmically scaled. Each bubble represents one dataset, with bubble area proportional to total recording hours. 
    LibriBrain100 occupies a distinctive position: it combines a competitive number of subjects with substantially more total recording time and substantially greater single-subject depth than most existing datasets. 
    The arrow indicates the expansion from LibriBrain to LibriBrain100.
    }
    \label{fig:dataset-comparison}
\end{figure}

Figure~\ref{fig:dataset-comparison} compares LibriBrain100 against other public MEG speech datasets along three dimensions: cohort breadth, single-subject depth, and total recording time. Each point corresponds to one dataset, with the x-axis showing the number of subjects and the y-axis showing the maximum number of recording hours available for any single subject. Bubble area is proportional to the total number of recording hours in the dataset. The arrow highlights the step from LibriBrain to LibriBrain100: LibriBrain100 substantially increases the total recording time and number of subjects while preserving the unusually deep single-subject regime that made LibriBrain useful for within-subject speech decoding. 
The public MEG speech datasets compared to LibriBrain100~\citep{mantegna2026libribrain100} are LibriBrain~\citep{ozdogan2025libribrain}, Armeni~\citep{armeni2022}, Le Petit Prince~\citep{dascoli2025natcomm}, MEG-MASC~\citep{gwilliams2023megmasc}, and MOUS~\citep{schoffelen2019mous}. 

\section{Retrieval Vocabulary}\label{sec:retrieval_vocab}

Table~\ref{tab:words} lists the two 50-word retrieval vocabularies used for evaluating word classification in this competition. Both metrics defined in Section~\ref{sec:metrics} (top-10 balanced accuracy and OVMI) are computed independently on each vocabulary and reported separately. Only $\mathrm{BAcc@10}$ on the competition vocabulary is the primary metric used for competition leaderboards and prize decisions.

\begin{table*}[t]
\centering
\small
\setlength{\tabcolsep}{15pt}

\textit{(a) Competition vocabulary (50 words)}\\[2pt]
\begin{tabular}{rl@{\hspace{0.8em}}rl@{\hspace{0.8em}}rl@{\hspace{0.8em}}rl@{\hspace{0.8em}}rl}
\toprule
\# & Word & \# & Word & \# & Word & \# & Word & \# & Word \\
\midrule
1  & a       & 11 & be   & 21 & him  & 31 & on     & 41 & they \\
2  & all     & 12 & but  & 22 & i    & 32 & our    & 42 & think \\
3  & always  & 13 & can  & 23 & in   & 33 & out    & 43 & this \\
4  & am      & 14 & do   & 24 & is   & 34 & people & 44 & time \\
5  & an      & 15 & for  & 25 & it   & 35 & really & 45 & to \\
6  & and     & 16 & good & 26 & it's & 36 & she    & 46 & very \\
7  & any     & 17 & had  & 27 & my   & 37 & so     & 47 & was \\
8  & are     & 18 & has  & 28 & new  & 38 & that   & 48 & we \\
9  & as      & 19 & have & 29 & not  & 39 & the    & 49 & were \\
10 & at      & 20 & he   & 30 & of   & 40 & there  & 50 & will \\
\bottomrule
\end{tabular}

\vspace{1em}

\textit{(b) Moses 50 vocabulary~\citep{moses2021nejm}}\\[2pt]
\begin{tabular}{rl@{\hspace{0.8em}}rl@{\hspace{0.8em}}rl@{\hspace{0.8em}}rl@{\hspace{0.8em}}rl}
\toprule
\# & Word & \# & Word & \# & Word & \# & Word & \# & Word \\
\midrule
1  & am          & 11 & faith   & 21 & here    & 31 & need    & 41 & that \\
2  & are         & 12 & family  & 22 & hope    & 32 & no      & 42 & they \\
3  & bad         & 13 & feel    & 23 & how     & 33 & not     & 43 & thirsty \\
4  & bring       & 14 & glasses & 24 & hungry  & 34 & nurse   & 44 & tired \\
5  & clean       & 15 & going   & 25 & i       & 35 & okay    & 45 & up \\
6  & closer      & 16 & good    & 26 & is      & 36 & outside & 46 & very \\
7  & comfortable & 17 & goodbye & 27 & it      & 37 & please  & 47 & what \\
8  & coming      & 18 & have    & 28 & like    & 38 & right   & 48 & where \\
9  & computer    & 19 & hello   & 29 & music   & 39 & success & 49 & yes \\
10 & do          & 20 & help    & 30 & my      & 40 & tell    & 50 & you \\
\bottomrule
\end{tabular}

\caption{The two 50-word retrieval vocabularies used in the word-classification task, listed alphabetically with index values for reference. (a) The \emph{competition vocabulary}, selected from words that have sufficient occurrences for low-variance evaluation. (b) The \emph{Moses 50} vocabulary~\citep{moses2021nejm}, developed for assistive communication and used in invasive decoding studies.}
\label{tab:words}
\end{table*}

\textbf{Competition vocabulary.} The first vocabulary is a 50-word set selected to achieve low-variance performance estimation. Two criteria guided selection. First, words were required to occur frequently enough to provide sufficient examples for both training and held-out evaluation. Second, the set was constructed to span multiple grammatical categories---function words (determiners, pronouns, auxiliaries, prepositions, conjunctions) together with common content words (e.g.\ \emph{time}, \emph{good}, \emph{think}, \emph{people}, \emph{new})---so that the vocabulary supports the composition of practically meaningful utterances. This design preserves continuity with recent non-invasive word-decoding work, where similar high-frequency vocabularies are standard~\citep{dascoli2025natcomm, jayalath2025meg-xl}. 

\textbf{Moses 50.} The second vocabulary is the 50-word list of~\citet{moses2021nejm}, developed with patient input for assistive communication and subsequently adopted in invasive decoding studies~\citep{willett2023high}. In contrast to the competition vocabulary, Moses 50 emphasises content words tied to clinical and daily-living needs (\emph{hungry}, \emph{thirsty}, \emph{tired}, \emph{glasses}, \emph{nurse}, \emph{music}) and social or affective expressions (\emph{hello}, \emph{goodbye}, \emph{please}, \emph{hope}). Reporting results on Moses 50 enables direct comparison between non-invasive systems evaluated here and invasive systems evaluated against the same vocabulary for the first time. But even with the size of the LibriBrain100 corpus, not all of the Moses 50 words are adequately attested in it. Evaluating on the Moses 50 data alone would underrepresent what is possible and the words in it are not the only words we want to decode.

Despite their modest size, both vocabularies support a useful range of short utterances. From Moses 50: assistive expressions such as ``i feel tired'', ``please bring my glasses'', ``i am not comfortable'', and ``tell my family''. From the competition vocabulary: short conversational forms such as ``i think so'', ``we have time'', ``i am very good'', ``i am not good at all'', ``can i have this'', and ``we can do this''. The two vocabularies share 13 words, so their union covers 87 distinct word types.

\section{Stimulus Materials}
\label{app:corpora}

Subject~0 listened to the complete \textit{Sherlock Holmes} canon \citep{doyle1887study,doyle1890sign,doyle1892adventures,doyle1893memoirs,doyle1902hound,doyle1905return,doyle1915valley,doyle1917lastbow,doyle1927casebook}, the TIMIT acoustic-phonetic corpus \citep{garofolo1993timit}, the MOCHA-TIMIT articulatory database \citep{wrench1999mocha_timit}, and 30 podcast stories from The Moth, a subset of those used in \citet{tang2023}. Subjects~1--32 listened to the validation and test recordings from LibriBrain \citep{ozdogan2025libribrain}. Further details for stimuli, recording protocols, and preprocessing can be found in the LibriBrain100 dataset paper \citep{mantegna2026libribrain100}.

\section{Code Samples}\label{sec:code_samples}

The recommended way to participate in the competition is through our custom Python library, \texttt{pnpl} (\verb|pip install pnpl|), which integrates ready-to-use PyTorch dataloaders, including automatic dataset downloads from Hugging Face, and task configurations:

\begin{minted}{python}
from pnpl.datasets import LibriBrain100
from pnpl.tasks import WordClassification

ds = LibriBrain100(
    data_path="./data/LibriBrain100",   # data downloaded if missing from path
    task=WordClassification(tmin=0.2, tmax=0.6),
    partition="train",
)

x, y = ds[0]   # x: (channels, time), y: integer word class id
\end{minted}

In addition, it includes a small utility for submitting predictions to the Kaggle competition.

\begin{minted}{python}
from pnpl.competition import write_submission, submit_to_kaggle
  
write_submission(
    "submission.csv",
    indices=indices,            # (N,)
    primary_probs=primary,      # (N, 50) — Competition Vocabulary
    secondary_probs=secondary,  # (N, 50) — Moses Vocabulary
)
submit_to_kaggle("submission.csv")
\end{minted}

\end{document}